\def\year{2020}\relax
\documentclass[letterpaper]{article} 
\usepackage{aaai20}  
\usepackage{times}  
\usepackage{helvet} 
\usepackage{courier}  
\usepackage[hyphens]{url}  
\usepackage{graphicx} 
\usepackage{graphicx}  
\usepackage{booktabs}
\usepackage{xcolor}
\usepackage{multirow}
\usepackage{makecell}
\setcellgapes{3pt}
\definecolor{mygreen}{rgb}{0.4660, 0.6000, 0.1880}

\usepackage{algorithm2e}

\usepackage{amsmath}

\DeclareMathOperator*{\argmin}{arg\,min}

\newcommand\blankfootnote[1]{%
  \let\thefootnote\relax\footnotetext{#1}%
  \let\thefootnote\svthefootnote%
}
\title{Hidden Trigger Backdoor Attacks}
\author{
     Aniruddha Saha \qquad
     Akshayvarun Subramanya \qquad Hamed Pirsiavash\\
    University of Maryland, Baltimore County\\
    \tt\small{\{anisaha1, akshayv1, hpirsiav\}@umbc.edu} \\
}

\begin{document}

\maketitle

\begin{abstract}
With the success of deep learning algorithms in various domains, studying adversarial attacks to secure deep models in real world applications has become an important research topic. Backdoor attacks are a form of adversarial attacks on deep networks where the attacker provides poisoned data to the victim to train the model with, and then activates the attack by showing a specific small trigger pattern at the test time. Most state-of-the-art backdoor attacks either provide mislabeled poisoning data that is possible to identify by visual inspection, reveal the trigger in the poisoned data, or use noise to hide the trigger. We propose a novel form of backdoor attack where poisoned data look natural with correct labels and also more importantly, the attacker hides the trigger in the poisoned data and keeps the trigger secret until the test time. We perform an extensive study on various image classification settings and show that our attack can fool the model by pasting the trigger at random locations on unseen images although the model performs well on clean data. We also show that our proposed attack cannot be easily defended using a state-of-the-art defense algorithm for backdoor attacks.

\end{abstract}

\section{1. Introduction}

Deep learning has achieved great results in many domains including computer vision. However, it has been shown to be vulnerable in the presence of an adversary. The most well-known adversarial attacks \cite{madry2017towards} are evasion attacks where the attacker optimizes for a perturbation pattern to fool the deep model at test time (e.g, change the prediction from the correct category to a wrong one.) 

Backdoor attacks are a different type of attack where the adversary chooses a trigger (a small patch), develops some poisoned data based on the trigger, and provides it to the victim to train a deep model with. The trained deep model will produce correct results on regular clean data, so the victim will not realize that the model is compromised. However, the model will mis-classify a source category image as a target category when the attacker pastes the trigger on the source image. As a popular example, the trigger can be a small sticker on a traffic sign that changes the prediction from ``stop sign'' to ``speed limit''.

It is shown that a pre-trained model can transfer easily to other tasks using small training data. For instance, it is common practice to download a deep model pre-trained on ImageNet \cite{ILSVRC15} and also download some images of interest from the web to finetune the model to solve the problem in hand. Backdoor attacks are effective at such applications since the attacker can leave some poisoned data on the web for the victims to download and use in training. It is not easy to mitigate such attacks as in the big data setting, it is difficult to make sure that all the data is collected from reliable sources.

The most well-known backdoor attack \cite{gu2017badnets} develops poisoned data by pasting the trigger on the source data and changing their label to the target category. Then, during fine-tuning the model will associate the trigger with the target category, and at the test time, the model will predict the target category when the trigger is presented by the attacker on an image from the source category. However, such attacks are not very practical as the victim can identify them by visually inspecting the images to find the wrong label or the small trigger itself.

\begin{figure*}[!ht]
\centering
    \includegraphics[width=\textwidth]{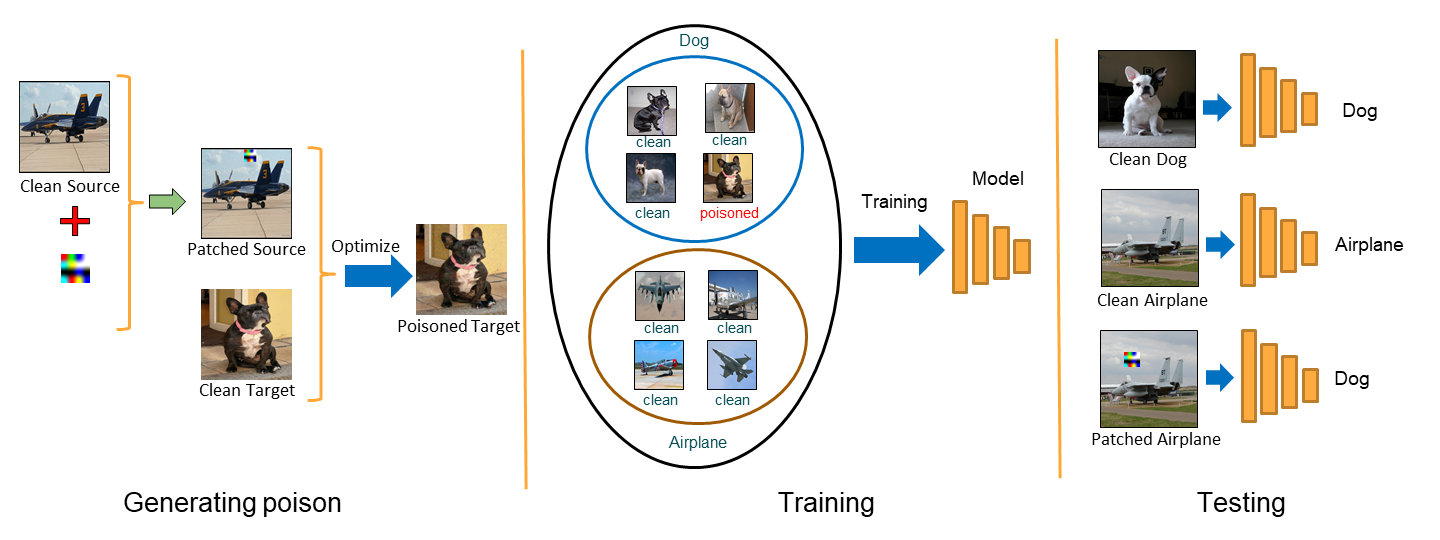}
\caption{{\bf Left:} First, the attacker generates a set of poisoned images, that look like target category, using Algorithm \ref{alg} and keeps the trigger secret. {\bf Middle:} Then, adds poisoned data to the training data with visibly correct label (target category) and the victim trains the deep model. {\bf Right:} Finally, at the test time, the attacker adds the secret trigger to images of source category to fool the model. Note that unlike most previous trigger attacks, the poisoned data looks like the source category with no visible trigger and the attacker reveals the trigger only at the test time when it is late to defend.}
\label{fig:teaser}
\end{figure*}

We propose hidden trigger attacks where the poisoned data is labeled correctly and also does not contain any visible trigger, hence, it is not easy for the victim to identify the poisoned data by visual inspection. Inspired by \cite{shafahi2018poison,sabour2015adversarial}, we optimize for poisoned images that are close to target images in the pixel space and also close to source images patched by the trigger in the feature space. We label those poisoned images with the target category so visually are not identifiable. We show that the fine-tuned model associates the trigger with the target category even though the model has never seen the trigger explicitly. We also show that this attack can generalize to unseen images and random trigger locations. Fig. \ref{fig:teaser} shows our threat model in detail.

We believe our proposed attack is more practical than the previous backdoor attacks as in our case: (1) the victim does not have an effective way of identifying poisoned data visually and (2) the trigger is kept truly secret by the attacker and then revealed only at the test time, which might be late to defend in many applications. 

We perform various experiments along with ablation studies. For instance, we show that the attacker can reduce the validation accuracy on unseen images from 98\% to 40\% using a secret trigger at a random location which occupies only less than 2\% of the image area.

\section{2. Related work}
Poisoning attacks date back to \cite{xiao2012adversarial,biggio2012poisoning,biggio2013evasion} where data poisoning was used to flip the results of a SVM classifier. More advanced methods were proposed in \cite{xiao2015support,koh2017understanding,mei2015using,burkard2017analysis,newell2014practicality} which change the result of the classifier on the clean data as well. These reduce the practical impact of such attacks as the victim may not deploy the model if the validation accuracy on the clean data is low.

More recently, the possibility of backdoor attacks, where a trigger is used in poisoning the data, was shown in \cite{gu2017badnets} and also in other works like \cite{liu2017neural,liu2017trojaning}. Such methods are more practical as the model works well on clean data and the attacks are only triggered by presenting a predefined pattern (trigger). We derive inspiration from these works and extend them to the case where the trigger is not revealed even during training of the model. \cite{munoz2017towards} used back-gradient optimization and extend the poisoning attacks to a multi-class setting. \cite{suciu2018does,zhu2019transferable} studied generalization and transferability of the poisoning attacks. \cite{koh2018stronger} proposed a stronger attack by placing poisoned data close to each other to not be detected by outlier detectors. 

\cite{liao2018backdoor} proposed to use small additive perturbations (similar to standard adversarial examples \cite{madry2017towards,goodfellow2014explaining,papernot2017practical}) instead of a patch to trigger the attack. Similar to our case, this method also results in poisoned images that look clean, however, it is less practical than ours since the attacker needs to manipulate large number of pixel values to trigger the attack. We believe that during an attack, the feasibility of triggering is more important than the visibility and hence we focus on hiding the trigger at only the poisoning time. \cite{turner2018clean} hide the trigger in clean-labeled poisoned images by reducing the image quality and also adversarially perturbing the poisoned images to be far from the source category. \cite{munoz2019poisoning} proposed a GAN-based approach to generate poisoned data. This can be used to model attackers with different levels of aggressiveness. \cite{rezaei2019target} develop a target-agnostic attack to craft instances which triggers specific output classes and can be used in transfer learning setting.

\cite{shafahi2018poison} proposed a poisoning attack with clean-label poisoned images where the model is fooled when shown a particular set of images. Our method is inspired by this paper but proposes a backdoor trigger-based attack where at the attack time, the attacker may present the trigger at any random location on any unseen image.

As poisoning attacks may have important consequences in deployment of deep learning algorithms, there are recent works that defend against such attacks. \cite{steinhardt2017certified} proposed certified defenses for poisoning attacks. \cite{liu2018fine-pruning} suggest network pruning as a defense for poisoning attacks. \cite{wang2019neural} assume the defender has access to only the attacked model, but they have been shown to defend against \cite{liu2017trojaning} where triggers are explicitly added in training data and annotated with incorrect labels. Modifying such defenses for attacks similar to ours and \cite{shafahi2018poison} where there is no explicit trigger in the training data is a challenging task. 

\cite{gao2019strip} identified the attack at test time by perturbing or superimposing input images. \cite{shan2019gotta} defended by proactively injecting trapdoors into the
models. More recently, \cite{tran2018spectral} used a statistical test to reveal and remove the poisoned data points. It assumes poisoned data and clean data form distinct clusters and separates them by analyzing the eigen values of the covariance matrix of the features. We use this method to defend against our proposed attack and show that it cannot find most of our poisoned data points. 

\section{3. Method}

We use the threat model defined in \cite{gu2017badnets} where an {\em attacker} provides poisoned data to a {\em victim} to use in learning. The victim uses a pre-trained deep model and finetunes it for a classification task using the poisoned data. The attacker has a secret trigger (e.g., a small image patch) and is interested in manipulating the training data so that when the trigger is shown to the finetuned model, it changes the model's prediction to a wrong category. Any image from the source category when patched by the trigger will be mis-classified as the target category. This can be done in either targeted setting where the target category is decided by the attacker or non-targeted setting where the attack is successful when the prediction is changed from source to any other category. Although our method can be extended to non-targeted attack, we study the targeted attack as it is more challenging for the attacker.

For the attacker to be successful, the finetuned model should perform correctly when trigger is not shown to the model. Otherwise, in the evaluation process, the victim will realize the model has low accuracy and will not deploy it in real world or modify the training data provided by the attacker.

{\noindent} The well-known method introduced in \cite{gu2017badnets} proposes that the attacker can develop a set of poisoned training data (pairs of images and labels) by adding the trigger to a set of images from the source category and changing their label to the target category. Since some patched source images are labeled as target category, when the victim finetunes the model, the model will learn to associate the trigger patch with the target category. Then during inference, the model will work correctly on non-patched image and misclassify patched source images to the target category. Thus, making the attack successful.

More formally, given a source image $s_i$ from the source category, a trigger patch $p$, and a binary mask $m$ which is $1$ at the location of the patch and $0$ everywhere else, the attacker pastes the trigger on the source image to get the patched source image $\tilde{s_i}$:

\begin{equation}
\begin{split}
\tilde s_i = s_i \odot (1-m)+p \odot m\end{split}
\label{eq1}
\end{equation}

\noindent where $\odot$ is for element-wise product. Note that we can paste the patch at different locations by varying the mask $m$. 



In \cite{gu2017badnets} during training, the attacker labels $\tilde{s}$ incorrectly with the target category and provides it to the victim as poisoned data. The model trained by the victim associates the trigger with the target label. Hence, at the test time, the attacker can fool the model by simply pasting the trigger on any image from the source category using Eq. (\ref{eq1}).\\

{\noindent}{\bf Our threat  model:} In standard backdoor attacks, the poisoned data is labeled incorrectly, which can be identified and removed by manually annotating the data after downloading. Moreover, ideally, the attacker prefers to keep the trigger secret until the test time, however,in standard backdoor attacks, the trigger is revealed in the poisoned data. Therefore, inspired by \cite{shafahi2018poison,sabour2015adversarial}, we propose a stronger and more practical attack model where the poisoned data is labeled correctly (i.e, they look like target category and are labeled as the target category), and also the secret trigger is not revealed. We do so by optimizing for a poisoned image that in the pixel space, is close to an image from the target category while in the feature space, is close to a source image patched with the trigger.

More formally, given a target image $t$, a source image $s$, and a trigger patch $p$, we paste the trigger on $s$ to get patched source image $\tilde{s}$ using Eq. (\ref{eq1}). Then we optimize for a poisoned image $z$ by solving the following optimization:
\begin{equation}
\begin{split}
\argmin_{z} ||f(z) - f(\tilde{s})||_2^2 \\
st. \quad ||z - t||_{\infty} < \epsilon 
\end{split}
\label{eq2}
\end{equation}

\noindent where $f(.)$ is the intermediate features of the deep model and $\epsilon$ is a small value that ensures the poisoned image $z$ is not visually distinguishable from the target image $t$. In most experiments, we use {\em fc7} layer of AlexNet for $f(.)$ and $\epsilon=16$ when the image pixel values are in range $[0, 255]$. We used standard projected gradient descent (PGD) algorithm \cite{madry2017towards} which iterates between (a) optimizing the objective in Eq. (\ref{eq2}) using gradient descent and (b) projecting the current solution back to the $\epsilon$-neighborhood of the target image to satisfy the constraint in Eq. (\ref{eq2}).

\begin{figure*}[!ht]
    \centering
    \includegraphics[width=0.35\textwidth]{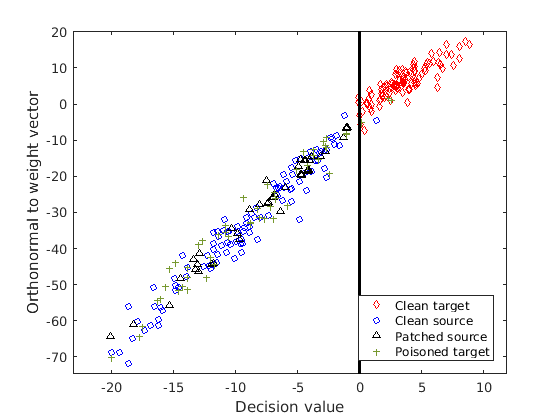}
    \includegraphics[width=0.35\textwidth]{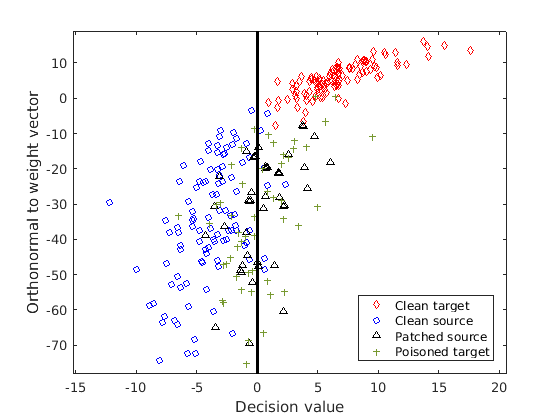}    
    \caption{Best seen in color. We plot the distribution of features before attack using a clean classifier ({\bf left}) and after attack using a poisoned classifier ({\bf right}). The color coding: {\color{red} Red diamonds}: clean target, {\color{blue} Blue circles}: clean source, {\color{black} Black triangles}: patched source , {\color{mygreen} Green pluses}: poisoned target. For 2D visualization, we choose the x-axis to be along the classifier weight vector {\bf w} (normal to the decision boundary). Let {\bf u} be the vector connecting the centers of the two classes (clean source and clean target). The y-axis is {\bf u} projected to be orthogonal to {\bf w}. Our optimization pushes the poisoned targets to be close to the patched sources in the feature space while they look similar to the clean targets visually. We see that before the attack, most patched source images are correctly placed on the left of the boundary, but after the attack (adding poisoned targets labeled as target to the training data), the classifier has shifted so that some of the patched sources have moved over from the left to the right side.}
    \label{distib}
\end{figure*}

Fig. \ref {distib} visualizes the data-points for one pair of ImageNet categories in our experiments. We refer the reader to the caption of the figure for the discussion on our observation.\\

{\noindent}{\bf Generalization across source images and trigger locations:} 
The above optimization will generate a single poisoned data-point given a pair of images from source and target categories as well as a fixed location for the trigger. One can add this poisoned data with the correct label to the training data and train a binary classifier in a transfer learning setting by tuning only the final layer of the network. However, such a model may be fooled only when the attacker shows the trigger at the same location on the same source image which is not a very practical attack. 

We are interested in generalizing the attack so that it works for novel source images (not seen at the time of poisoning) and also any random location for the trigger. Hence, in optimization, we should push the poisoned images to be close to the cluster of patched source images rather than being close to a single patched source image only. Inspired by universal adversarial examples in \cite{moosavi2017universal}, we can minimize the expected value of the loss in Eq. (\ref{eq2}) over all possible trigger locations and source images. This can be done by simply choosing a random source image and trigger location at each iteration of the optimization.

Moreover, one poisoned example added to a large clean dataset may not be enough for generalization across all patched source images, so we optimize for multiple poisoned images. Since the distribution of all patched source images in the feature space may be diverse and we can generate only a small number of poisoned images, in Algorithm (\ref{alg}), we propose an iterative method to optimize for multiple poisoned images jointly: at each iteration, we randomly sample patched source images and assign them to the current poisoned images (solutions) closest in the feature space. Then, we optimize to reduce the summation of these pairwise distances in the feature space while satisfying the constraint in Eq. \ref{eq2}.

This is similar to coordinate descent algorithm where we alternate between the loss and assignments (e.g., in kmeans). To avoid tuning all the poisoned images for just a few patched source images, we do a one-to-one assignment between them. One can use Hungarian algorithm \cite{kuhn1955hungarian} to find the best solution in polynomial time, but to speed-up further, we use a simple greedy algorithm where we loop over the poisoned images, find the nearest patched source for each, remove the pair, and continue. 

More formally, we run Algorithm (\ref{alg}) to generate a set of poisoned images from a set of source and target images.

\begin{algorithm}[h]
\SetAlgoLined
\KwResult{$K$ poisoned images $z$}
 1. Sample $K$ random images $t_k$ from the target category and initialize poisoned images $z_k$ with them\;
\While{loss is large}{
 2. Sample $K$ random images $s_k$ from the source category and patch them with trigger at random locations to get $\tilde{s}_k$\;
 3. Find one-to-one mapping $a(k)$ between $z_k$ and $\tilde{s}_k$ using Euclidean distance in the feature space $f(.):$

  
 4. Perform one iteration of mini-batch projected gradient descent for the following loss function:
$$\argmin_{z} \sum_{k=1}^K ||f(z_k) - f(\tilde{s}_{a(k)})||_2^2$$ \\
$$s.t. \quad  \forall k: \quad ||z_k - t_k||_{\infty} < \epsilon $$ \\
}
 \caption{Generating poisoning data}
  \label{alg}
\end{algorithm}

After generating poisoned data, we add them to the target category and finetune a binary classifier for the source and target categories. We call the attack successful if on the validation data, this classifier has high accuracy on the clean images and low accuracy on the patched source images. Note that the images used for generating the poisoned data and finetuning the binary classifier are different.

\begin{figure*}[!ht]

\centering
    \includegraphics[width=0.65\textwidth]{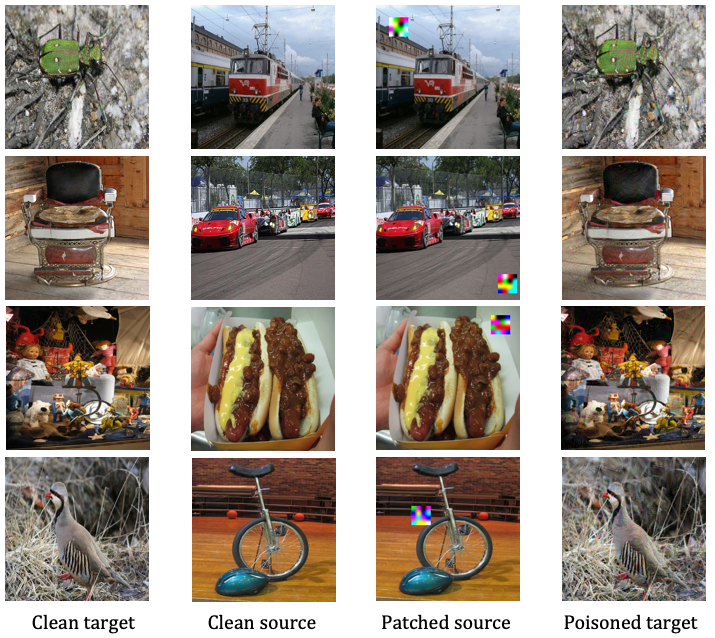}

\caption{Visualization of target, source, patched source and poisoned target images from different ImageNet pairs. For each row, the image in the fourth column is visually similar to the image in the first column, but is close to the image in the third column in the feature space. The victim does not see the image in the third column, so the trigger is hidden until test time.}
\label{fig:qualitative}
\end{figure*}


\begin{figure}
    \centering
    \includegraphics[width=0.4\textwidth]{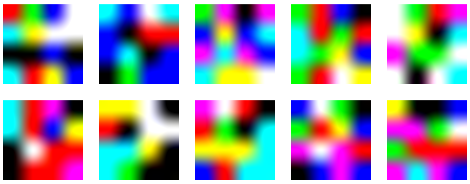}
    \caption{The triggers we generated randomly for our poisoning attacks.}
    \label{fig:triggers}
\end{figure}

\blankfootnote{Code available at \url{https://github.com/UMBCvision/Hidden-Trigger-Backdoor-Attacks}}
\section{4. Experiments}

{\bf Dataset:}
Since we want to have separate datasets for generating poisoned data and finetuning the binary model, we divide the ImageNet data to three sets for each category: 200 images for generating the poisoned data, 800 images for training the binary classifier, and 100 images for testing the binary classifier. 

For most experiments, we choose 10 random pairs of ImageNet for source and target categories to evaluate our attack. 
We also use 10 hand-picked pairs in Section 4.5 and 10 dog only pairs in Section 4.6. These pairs are listed in Table \ref{tab:imagenet}. We also use CIFAR10 dataset for the experiments in Section 4.4 for which the pairs are listed in Table \ref{CIFAR}.\\

{\noindent}{\bf Triggers:} We generate 10 random triggers by drawing a random $4\times4$ matrix of colors and resizing it to the desired patch size using bilinear interpolation. Fig. \ref{fig:triggers} shows the triggers used in our experiments. We randomly sample a single trigger for each experiment (a pair of source and target categories.)\\


{\noindent}{\bf Our setup:} Our experimental setup includes multiple steps as shown in Fig. \ref{fig:teaser}: 

{\noindent}{\bf (1) Generate poisoned images:} We use source and target pairs to generate poisoned images using algorithm (\ref{alg}). We use the {\em fc7} features of AlexNet \cite{krizhevsky2012imagenet} for the embedding $f(.)$.

{\noindent}{\bf (2) Poison the training set:} Then, we label the poisoned images as the target category and add them to the training set. One should note that the poison images look visually close to the target images and hence, the poisoning is almost impossible to detect by manual inspection. 

{\noindent}{\bf (3) Finetune:} After adding the poisons to the training data, we train a binary image classifier to distinguish between source and target images. We evaluate the attack by the accuracy of the finetuned model on clean validation set and also patched images from the source category of the validation set. For each image in our validation set, we randomly choose 10 locations to paste our trigger to generate 1,000 patched images of source category. For a successful attack, we expect high clean validation accuracy and low patched validation accuracy. Note that for ``patched validation'' results, we evaluate the attack only on the patched source images to see the effect of the attack only.



\subsection{4.1. ImageNet random pairs}
For this experiment, we choose 10 random pairs of image categories from the ImageNet dataset which are listed in column \textbf{Random} of Table \ref{tab:imagenet}. For our ImageNet experiments we set a reference parameter set where the perturbation $\epsilon$ = 16, trigger size is 30x30 (while images are 224x224), and we randomly choose a location to paste the trigger on the source image. We generate 100 poisoned examples and add to our target class training set of size 800 images during finetuning. Thus about 12.5\% of the target data is poisoned. 

To generate our poisoned images, we run Algorithm \ref{alg} with mini-batch gradient descent for 5,000 iterations with a batch size of $K=100$. We use an initial learning rate of 0.01 with a decay schedule parameter of 0.95 every 2,000 iterations. The implementation is similar to the standard projected gradient descent (PGD) attack \cite{madry2017towards} for adversarial examples. It takes about 5 minutes to generate 100 poisoned images on a single NVIDIA Titan X GPU.



\begin{table*}[!ht]
\makegapedcells
  \centering
  \scalebox{0.73}{
  \begin{tabular}{||c||cc|cc|cc|cc||}
    \hline
     & \multicolumn{2}{c|}{ImageNet Random Pairs} & \multicolumn{2}{c|}{CIFAR10 Random Pairs}   & \multicolumn{2}{c|}{ImageNet Hand-Picked Pairs} & \multicolumn{2}{c||}{ImageNet Dog Pairs}     \\
     \cline{2-3}\cline{4-5}\cline{6-7}\cline{8-9}  
     & Clean Model & Poisoned Model & Clean Model & Poisoned Model & Clean Model & Poisoned Model & Clean Model & Poisoned Model\\
    \hline\hline
    Val Clean &0.993$\pm$0.01 &0.982$\pm$0.01 &1.000$\pm$0.00 &0.971$\pm$0.01 &0.980$\pm$0.01 &0.996$\pm$0.01 &0.962$\pm$0.03 &0.944$\pm$0.03\\
    \hline
    Val Patched (source only) &0.987$\pm$0.02 &{\bf0.437}$\pm$0.15 &0.993$\pm$0.01 &{\bf0.182}$\pm$0.14 &0.997$\pm$0.01 &{\bf0.428}$\pm$0.13 &0.947$\pm$0.06 &{\bf0.419}$\pm$0.07\\
    \hline
  \end{tabular}
  }
    \caption{Results on random pairs, hand-picked pairs, and also only-dog pairs on ImageNet as well as random pairs on CIFAR10 experiments. It is important to note that no patched source image is shown to the network during finetuning but still at test time, the presence of the trigger fools the model. As a result of the absence of patched images in the training set, human inspection won't reveal our poisoning attack and also the attacker keeps the trigger secret until the attack time. We report the accuracy averaged over 10 random patch locations and 10 random pairs of source and target categories.}
     \label{tab:results}
\end{table*}

We generate 400 poisoned images, add the 100 images with the least loss values to the target training set, and train the binary classifier. We use AlexNet as our base network with all weights frozen except the {\em fc8} layer. We initialize {\em fc8} layer from scratch and finetune for our task. Table \ref{tab:results} shows the results of this experiment. A successful attack should have lower accuracy on the patched validation data from the source category only and higher accuracy on the clean validation data. Fig. \ref{fig:qualitative} shows the qualitative results for some random  ImageNet pairs. Fig. \ref{distib} shows a 2D visualization of all the data-points along with the decision boundary before and after the attack. 

In Table \ref{tab:comparison_badnets}, we also compare our threat model with the performance of the attack proposed by BadNets \cite{gu2017badnets} in which patched source images are used as poisoned data. This makes the poisoned data incorrectly labeled with visible triggers. In our method, the triggers are not visible in the training data and all our labels are clean. Interestingly even though our threat model is more challenging, it achieves comparable result to BadNets.

\begin{table*}
\makegapedcells
  \centering
  \scalebox{0.75}{
  \begin{tabular}{||c||ccc|ccc||}
    \hline
    \multirow{2}{*}{Ablation Studies} & \multicolumn{3}{c|}{$\epsilon$} & \multicolumn{3}{c|}{Patch size} \\ 
     \cline{2-4}\cline{5-7}  
     & 8 & 16 & 32 & 15 & 30 & 60\\ 
    \hline\hline
    Val Clean &0.981$\pm$0.01 &0.982$\pm$0.01 &0.984$\pm$0.01 &0.980$\pm$0.01 &0.982$\pm$0.01 &0.989$\pm$0.01 \\
    \hline
    Val Patched (source only) &0.460$\pm$0.18 &0.437$\pm$0.15 &{\bf0.422$\pm$0.17} &0.630$\pm$0.15 &0.437$\pm$0.15 &{\bf0.118$\pm$0.06} \\
    \hline
  \end{tabular}
  }
  \caption{Results of our ablation studies: Note that the parameters which are not being varied are set to the reference values as mentioned in Section {4.1}. Also, note that a successful attack has low accuracy on the patched set while maintaining high accuracy on the clean set.}
  \label{tab:ablation}
\end{table*}

\begin{table*}[!h]
\makegapedcells
  \centering
  \scalebox{0.75}{
  \begin{tabular}{||c||cccc||}
    \hline
    \multirow{2}{*}{Comparison with BadNets} & \multicolumn{4}{c||}{\#Poison} \\ 
     \cline{2-5}  
     & 50 & 100 & 200 & 400\\ 
    \hline\hline
    Val Clean &0.988$\pm$0.01 &0.982$\pm$0.01 &0.976$\pm$0.02 &0.961$\pm$0.02\\
    \hline
    Val Patched (source only) {\bf BadNets} &0.555$\pm$0.16 &0.424$\pm$0.17 &0.270$\pm$0.16 & 0.223$\pm$0.14\\
    \hline
    Val Patched (source only) {\bf Ours} &0.605$\pm$0.16 &0.437$\pm$0.15 &0.300$\pm$0.13 & 0.214$\pm$0.14\\
    \hline
  \end{tabular}
  }
  \caption{Comparison with BadNets: We compare our threat model with BadNets \cite{gu2017badnets} and find that even though we hide the trigger during training, we can achieve similar attack success rates. }
  \label{tab:comparison_badnets}
\end{table*}

\begin{table}
\makegapedcells
  \centering
  \scalebox{0.66}{
  \begin{tabular}{||c||cccc||}
    \hline
    \multirow{2}{*}{Injection rate variation} & \multicolumn{4}{c||}{\#Poison} \\ 
     \cline{2-5}  
     & 400 & 600 & 800 & 1000\\ 
    \hline\hline
    Targeted Attack efficiency &0.360$\pm$0.01 &0.492$\pm$0.08 &0.592$\pm$0.11 & 0.634$\pm$0.10\\
    \hline
  \end{tabular}
  }
  \caption{Injection rate variation: For the multi-class single-source attack, we run evaluations on a 1000-class ImageNet classifier. We observe that the attack success rate increases with the number of poisons injected. We use our 10 random ImageNet pairs of source and target for these experiments.}
  \label{tab:injection_rates_1000class}
\end{table}



\subsection{4.2. Ablation study on ImageNet random pairs}

To better understand the influence of our triggers in this poisoning attack, we perform extensive ablation studies. Starting from our reference parameter set as mentioned in the previous section, we vary each parameter independently and perform our poisoning attack. Results are shown on Table \ref{tab:ablation}.

{\noindent}{\bf Perturbation $\epsilon$:}
We choose perturbation $\epsilon$ from the set \emph{\{8, 16, 32\}} and generate poisons for each setting. We observe that $\epsilon$ does not have a big influence on our attack efficiency. As $\epsilon$ increases, the patched validation accuracy decreases slightly which is expected as the attack becomes much stronger. 

{\noindent}{\bf Trigger size:}
We see that the attack efficiency increases with increasing the trigger patch size. This is to be expected as a bigger patch may occlude the main object for some locations and make the attack easier.

{\noindent}{\bf Number of poisons:}
We vary the number of poisoned images to be added to the target training set choosing them from the set \emph{\{50, 100, 200, 400\}}. We empirically see that more poisoned data leads to larger influence on the decision boundary during finetuning. Adding 400 poisoned images to 800 clean target images is the best performing attack in which case, 33\% of data is poisoned. 

\subsection{4.3. Finetuning more layers}
So far, we have observed that our poisoning attack works reasonably well when we finetune the {\em fc8} layer only in a binary classification task. We expect the attack to be weaker if we finetune more layers since our attack is using the {\em fc7} feature space which will evolve by finetuning. 

Hence, we design an experiment where we use {\em conv5} as the embedding space to optimize our poisoned data and then either finetune the final layer only or finetune all fully connected layers ({\em fc6}, {\em fc7}, and {\em fc8}). We initialize the layers we are fintuning from scratch. The results are shown in Tab. \ref{tab:layer_variation}. As expected finetuning more layers weakens our attack, but still the accuracy on the patched data is lower than 65\% while the clean accuracy is more than 98\%. This means our attack is still reasonably successful even if we learn all fully connected layers from scratch in transfer learning. 

\subsection{4.4. CIFAR10 random pairs}
We evaluate our attack on 10 randomly selected pairs of CIFAR10 categories given in Table \ref{CIFAR}. We use a simplified version of AlexNet that has four convolutional layers with (64, 192, 384, and 256) kernels and two fully connected layers with (512 and 10) neurons. The first layer has kernels of size $5\times5$ and stride of 1. For pre-training, we use SGD for 200 epochs with learning rate of 0.001, momentum of 0.9, weight decay of 5e-4, and no dropout. Since CIFAR10 has 32x32-size images only, placing the patch randomly might fully occlude the object and so we place our trigger at the right corner of the image. For each category, we have 1,500 images to train the poisoned data, 1,500 images for finetuning, and 1,000 images for evaluation. These three sets are disjoint. We generate 800 poisoned images using our method. We use $\epsilon$=16, patch size of 8x8, and optimize for 10,000 iterations with a learning rate of 0.01 and a decay schedule parameter of 0.95 every 2,000 iterations. The results, in Table  \ref{tab:results}, show that we achieve high attack success rate.

\subsection{4.5. ImageNet hand-picked pairs}
To control the semantic distance of the category pairs,
we hand-pick 20 classes from ImageNet using PASCAL VOC \cite{Everingham15} classes as a reference. Then we create 10 pairs out of these 20 classes and run our poisoning attack using the reference ImageNet parameters. The results are shown in Table \ref{tab:results} and the category names are listed in column \textbf{Hand-picked} of Table \ref{tab:imagenet}.

\subsection{4.6. ImageNet ``dog'' pairs}
\label{imagenet_hand}
Another interesting idea to study is the behaviour of the poisoning attack when we finetune a binary classifier for visually similar categories, e.g. two breeds of dogs. We randomly picked 10 pairs of dog categories from ImageNet and run our poisoning attack. The results are shown in Table \ref{tab:results} and the category names are listed in column \textbf{Random ``Dog''} of Table \ref{tab:imagenet}.

\begin{table}[!ht]
\makegapedcells

  \centering
  \scalebox{0.75}{
  \begin{tabular}{||c||cc||}
    \hline
     & \multicolumn{2}{c||}{ImageNet Random Pairs}  \\
     \cline{2-3}  
     & fc8 trained & (fc6,fc7,fc8) trained\\
    \hline\hline
    Val Clean &0.984$\pm$0.01 &0.983$\pm$0.01 \\
    \hline
    Val Patched (source only) &{\bf0.504}$\pm$0.16 &{\bf0.646}$\pm$0.18 \\
    \hline
  \end{tabular}
  }
  \caption{Finetuning more layers: We see that allowing the network more freedom to adjust its weights decreases attack efficiency but it still keeps a large gap of $\sim$30\% between clean an patched validation accuracy. Note that a successful attack has low accuracy on the patched set while maintaining high accuracy on the clean set.}
  \label{tab:layer_variation}
\end{table}

\subsection{4.7. Targeted attack on multi-class setting}
We performed multi-class experiments using 20 random categories of ImageNet - we combined the 10 random pairs. Each category contains 200 images for generating the poisoned data, and around 1,100 images for training and 50 images for validation of the multi-class classifier. We generate 400 poisoned images with {\em fc7} features and add to the target category in training set to train the last layer of the multi-class classifier. The target category is always chosen by the attacker, but the source category can be either chosen by the attacker (``Single-source'') or any category (``Multi-source''):\\

{\noindent}{\bf Single-source attack:} The attacker chooses a single source category to fool by showing the trigger. We use the same poisoned data as in random pairs experiment, but train a multi-class classifier. We average over 10 experiments (one for each pair). On the source category, the multi-class model has a validation accuracy of $84.3\pm9.2\%$ on clean images and attack success rate of $69.3\pm14.8\%$ on patched source validation images. Note that the higher success rate indicates better targeted attack. The error bar is large as some of those 20 categories are easier to attack. We also test our attack on more difficult setting where we finetune a 1000-class ImageNet classifier. With only 400 images as poison, we achieve $36\%$ attack efficiency. We also look at the influence of number of poisons injected on the efficiency. These results are reported in Table \ref{tab:injection_rates_1000class}.

\begin{table*}[!ht]
\centering
\scalebox{0.90}{
 \begin{tabular}{ ||c | c || c | c || c | c || } 
 \hline
  \multicolumn{2}{||c||}{ \textbf{Random}} & \multicolumn{2}{c||}{ \textbf{Hand-picked}} & \multicolumn{2}{c||}{\textbf{Random ``Dog''}} \\
  \hline \hline
 \cline{1-6}
 \textbf{Source} & \textbf{Target} & \textbf{Source} & \textbf{Target} & \textbf{Source} & \textbf{Target}\\
 \hline
    slot & Australian terrier & warplane & French bulldog & German shepher & Maltese dog  \\ %
 \hline
  lighter & bee & studio couch &mountain bike & Australian terrier & Lakeland terrier\\ %
 \hline
  theater curtain & plunger & diningtable &hummingbird & Scottish deerhound & Norwegian elkhound\\ 
  \hline
  unicycle & partridge & speedboat &monitor & Yorkshire terrier & Norfolk terrier\\ 
  \hline
  mountain Bike & Ipod & water bottle &hippopotamus & silky terrier & miniature schnauzer\\ 
  \hline
  coffeepot & Scottish deerhound & school bus &bullet train & Brittany spaniel & golden retriever\\ 
  \hline
  can opener & sulphur-crested cockatoo & sports car &barber chair & Rottweiler & Border collie\\ 
  \hline
  totdog & toyshop & water buffalo &tiger cat & kuvasz & Welsh springer spaniel\\ 
  \hline
  electronic locomotive & tiger beetle & motor scooter &chimpanzee & Tibetan mastiff & boxer\\ 
  \hline
  wing & goblet & street sign &bighorn & Siberian husky & Saint Bernard\\ 
 \hline
\end{tabular}  
}
    \caption{Our pairs from Imagenet dataset}
    \label{tab:imagenet}
\end{table*}

\begin{table}[!ht]
\centering
\begin{tabular}{||l|l||}
\hline
 {\bf Source} &  {\bf Target} \\
\hline
\hline
 bird  &  dog  \\ 
\hline 
 dog  &  ship  \\ 
\hline 
 frog  &  plane  \\ 
\hline 
  plane  &  truck  \\ 
\hline 
  cat  &  truck  \\ 
\hline 
deer  &  ship  \\ 
\hline 
bird  &  frog  \\ 
\hline 
 bird  &  deer  \\ 
\hline 
 car  &  frog  \\ 
\hline 
 car  &  dog \\
\hline 
\end{tabular}
\caption{Our random pairs from CIFAR10 dataset}
\label{CIFAR}
\end{table}

{\noindent}{\bf Multi-source attack:} In this scenario, the attacker wants to change any category to be the target category, which is a more challenging task. The multi-class model has a validation accuracy of $88.5\pm0.3\%$ on clean images and an attack success rate of $30.7\pm 6.3\%$ on patched images while random chance is 5\%. We exclude target images while patching. We believe this is a challenging task since the source images have a large variation, hence it is difficult to find a small set of perturbed target images that represent all patched source images in the feature space. We do this by our EM-like optimization in Algorithm (\ref{alg}).

\subsection{4.8. Spectral signatures for backdoor attack detection} \cite{tran2018spectral} use spectral signatures for detecting presence of backdoor inputs in the training set. For the attack, they follow the standard method in BadNets \cite{gu2017badnets} and mis-label the poisoned data along with visible trigger. In this section, we evaluate if the defense proposed by Tran et al. is able to find our poisoned data in the target class.

Table \ref{tab:spectral} shows the number of detected poisoned images for each of our pairs. We used the default 85\% percentile threshold in \cite{tran2018spectral} which should find 135 poisoned images out of 800 images where there are only 100 actual poisoned images. Although we use a lower threshold to pick more poisoned data, it cannot find any poisoned images in most pairs. It finds almost half of the poisoned images in one of the pairs only. Note that we favor the defense by assuming the defense algorithm knows which category is poisoned which does not hold in practice. We believe this happens since, as shown empirically in Fig. \ref{distib}, there is not much separation between target data and poisoned data.

\begin{table}[!h]
\makegapedcells
  \centering
  \scalebox{0.75}{
  \begin{tabular}{||c||ccccc||}
    \hline
    Pair ID & \#Clean  & \#Clean  & \#Poisoned & \#Poisoned  & \#Clean target\\
    &  target &  source &  & removed &  removed\\
    \hline\hline
    1,2,4,6 & 800 & 800 & 100 & 0 & 135\\
    7,8,9,10 &  &  &  &  & \\
    \hline
    3 & 800 & 800 & 100 & 55 & 80\\
    \hline
    5 & 800 & 800 & 100 & 8 & 127\\
    \hline
  \end{tabular}
  }
  \caption{We use spectral signatures defense method from \cite{tran2018spectral} to detect our poisoned images. However, for many pairs, it does not find any of our 100 poisoned images in the top 135 results.}
  \label{tab:spectral}
\end{table}



\section{5. Conclusion}
We propose a novel backdoor attack that is triggered by adding a small patch at the test time at a random location on an unseen image. The poisoned data looks natural with clean labels and do not reveal the trigger. Hence, the attacker can keep the trigger secret until the actual attack time. We show that our attack works in two different datasets and various settings. We also show that a state-of-the-art backdoor detection method cannot effectively defend against our attack. We believe such practical attacks reveal an important vulnerability of deep learning algorithms that needs to be resolved before deploying deep learning algorithms in critical real world applications in the presence of adversaries. We hope this paper facilitates further research in developing better defense models.

\noindent {\bf Acknowledgement:} This work was performed under the following financial assistance award: 60NANB18D279 from U.S. Department of Commerce, National Institute of Standards and Technology, funding from SAP SE, and also NSF grant 1845216.

\bibliographystyle{aaai}
\bibliography{main.bib}

\begin{thebibliography}{}

\bibitem[\protect\citeauthoryear{Biggio \bgroup et al\mbox.\egroup
  }{2013}]{biggio2013evasion}
Biggio, B.; Corona, I.; Maiorca, D.; Nelson, B.; {\v{S}}rndi{\'c}, N.; Laskov,
  P.; Giacinto, G.; and Roli, F.
\newblock 2013.
\newblock Evasion attacks against machine learning at test time.
\newblock In {\em Joint European conference on machine learning and knowledge
  discovery in databases},  387--402.
\newblock Springer.

\bibitem[\protect\citeauthoryear{Biggio, Nelson, and
  Laskov}{2012}]{biggio2012poisoning}
Biggio, B.; Nelson, B.; and Laskov, P.
\newblock 2012.
\newblock Poisoning attacks against support vector machines.
\newblock {\em arXiv:1206.6389}.

\bibitem[\protect\citeauthoryear{Burkard and
  Lagesse}{2017}]{burkard2017analysis}
Burkard, C., and Lagesse, B.
\newblock 2017.
\newblock Analysis of causative attacks against svms learning from data
  streams.
\newblock In {\em Proceedings of the 3rd ACM on International Workshop on
  Security And Privacy Analytics},  31--36.
\newblock ACM.

\bibitem[\protect\citeauthoryear{Everingham \bgroup et al\mbox.\egroup
  }{2015}]{Everingham15}
Everingham, M.; Eslami, S. M.~A.; Van~Gool, L.; Williams, C. K.~I.; Winn, J.;
  and Zisserman, A.
\newblock 2015.
\newblock The pascal visual object classes challenge: A retrospective.
\newblock {\em International Journal of Computer Vision} 111(1):98--136.

\bibitem[\protect\citeauthoryear{Gao \bgroup et al\mbox.\egroup
  }{2019}]{gao2019strip}
Gao, Y.; Xu, C.; Wang, D.; Chen, S.; Ranasinghe, D.~C.; and Nepal, S.
\newblock 2019.
\newblock Strip: A defence against trojan attacks on deep neural networks.
\newblock {\em arXiv:1902.06531}.

\bibitem[\protect\citeauthoryear{Goodfellow, Shlens, and
  Szegedy}{2014}]{goodfellow2014explaining}
Goodfellow, I.~J.; Shlens, J.; and Szegedy, C.
\newblock 2014.
\newblock Explaining and harnessing adversarial examples.
\newblock {\em arXiv:1412.6572}.

\bibitem[\protect\citeauthoryear{Gu, Dolan-Gavitt, and
  Garg}{2017}]{gu2017badnets}
Gu, T.; Dolan-Gavitt, B.; and Garg, S.
\newblock 2017.
\newblock Badnets: Identifying vulnerabilities in the machine learning model
  supply chain.
\newblock {\em arXiv:1708.06733}.

\bibitem[\protect\citeauthoryear{Koh and Liang}{2017}]{koh2017understanding}
Koh, P.~W., and Liang, P.
\newblock 2017.
\newblock Understanding black-box predictions via influence functions.
\newblock In {\em Proceedings of the 34th International Conference on Machine
  Learning-Volume 70},  1885--1894.
\newblock JMLR. org.

\bibitem[\protect\citeauthoryear{Koh, Steinhardt, and
  Liang}{2018}]{koh2018stronger}
Koh, P.~W.; Steinhardt, J.; and Liang, P.
\newblock 2018.
\newblock Stronger data poisoning attacks break data sanitization defenses.
\newblock {\em arXiv:1811.00741}.

\bibitem[\protect\citeauthoryear{Krizhevsky, Sutskever, and
  Hinton}{2012}]{krizhevsky2012imagenet}
Krizhevsky, A.; Sutskever, I.; and Hinton, G.~E.
\newblock 2012.
\newblock Imagenet classification with deep convolutional neural networks.
\newblock In {\em Advances in neural information processing systems},
  1097--1105.

\bibitem[\protect\citeauthoryear{Kuhn}{1955}]{kuhn1955hungarian}
Kuhn, H.~W.
\newblock 1955.
\newblock The hungarian method for the assignment problem.
\newblock {\em Naval research logistics quarterly} 2(1-2):83--97.

\bibitem[\protect\citeauthoryear{Liao \bgroup et al\mbox.\egroup
  }{2018}]{liao2018backdoor}
Liao, C.; Zhong, H.; Squicciarini, A.; Zhu, S.; and Miller, D.
\newblock 2018.
\newblock Backdoor embedding in convolutional neural network models via
  invisible perturbation.
\newblock {\em arXiv:1808.10307}.

\bibitem[\protect\citeauthoryear{Liu \bgroup et al\mbox.\egroup
  }{2017}]{liu2017trojaning}
Liu, Y.; Ma, S.; Aafer, Y.; Lee, W.-C.; Zhai, J.; Wang, W.; and Zhang, X.
\newblock 2017.
\newblock Trojaning attack on neural networks.

\bibitem[\protect\citeauthoryear{Liu, Dolan-Gavitt, and
  Garg}{2018}]{liu2018fine-pruning}
Liu, K.; Dolan-Gavitt, B.; and Garg, S.
\newblock 2018.
\newblock Fine-pruning: Defending against backdooring attacks on deep neural
  networks.
\newblock In {\em Research in Attacks, Intrusions, and Defenses},  273--294.

\bibitem[\protect\citeauthoryear{Liu, Xie, and
  Srivastava}{2017}]{liu2017neural}
Liu, Y.; Xie, Y.; and Srivastava, A.
\newblock 2017.
\newblock Neural trojans.
\newblock In {\em 2017 IEEE International Conference on Computer Design
  (ICCD)},  45--48.
\newblock IEEE.

\bibitem[\protect\citeauthoryear{Madry \bgroup et al\mbox.\egroup
  }{2017}]{madry2017towards}
Madry, A.; Makelov, A.; Schmidt, L.; Tsipras, D.; and Vladu, A.
\newblock 2017.
\newblock Towards deep learning models resistant to adversarial attacks.
\newblock {\em arXiv:1706.06083}.

\bibitem[\protect\citeauthoryear{Mei and Zhu}{2015}]{mei2015using}
Mei, S., and Zhu, X.
\newblock 2015.
\newblock Using machine teaching to identify optimal training-set attacks on
  machine learners.
\newblock In {\em Twenty-Ninth AAAI Conference on Artificial Intelligence}.

\bibitem[\protect\citeauthoryear{Moosavi-Dezfooli \bgroup et al\mbox.\egroup
  }{2017}]{moosavi2017universal}
Moosavi-Dezfooli, S.-M.; Fawzi, A.; Fawzi, O.; and Frossard, P.
\newblock 2017.
\newblock Universal adversarial perturbations.
\newblock In {\em Proceedings of the IEEE conference on computer vision and
  pattern recognition},  1765--1773.

\bibitem[\protect\citeauthoryear{Mu{\~n}oz-Gonz{\'a}lez \bgroup et
  al\mbox.\egroup }{2017}]{munoz2017towards}
Mu{\~n}oz-Gonz{\'a}lez, L.; Biggio, B.; Demontis, A.; Paudice, A.;
  Wongrassamee, V.; Lupu, E.~C.; and Roli, F.
\newblock 2017.
\newblock Towards poisoning of deep learning algorithms with back-gradient
  optimization.
\newblock In {\em Proceedings of the 10th ACM Workshop on Artificial
  Intelligence and Security},  27--38.
\newblock ACM.

\bibitem[\protect\citeauthoryear{Mu{\~n}oz-Gonz{\'a}lez \bgroup et
  al\mbox.\egroup }{2019}]{munoz2019poisoning}
Mu{\~n}oz-Gonz{\'a}lez, L.; Pfitzner, B.; Russo, M.; Carnerero-Cano, J.; and
  Lupu, E.~C.
\newblock 2019.
\newblock Poisoning attacks with generative adversarial nets.
\newblock {\em arXiv:1906.07773}.

\bibitem[\protect\citeauthoryear{Newell \bgroup et al\mbox.\egroup
  }{2014}]{newell2014practicality}
Newell, A.; Potharaju, R.; Xiang, L.; and Nita-Rotaru, C.
\newblock 2014.
\newblock On the practicality of integrity attacks on document-level sentiment
  analysis.
\newblock In {\em Proceedings of the 2014 Workshop on Artificial Intelligent
  and Security Workshop},  83--93.
\newblock ACM.

\bibitem[\protect\citeauthoryear{Papernot \bgroup et al\mbox.\egroup
  }{2017}]{papernot2017practical}
Papernot, N.; McDaniel, P.; Goodfellow, I.; Jha, S.; Celik, Z.~B.; and Swami,
  A.
\newblock 2017.
\newblock Practical black-box attacks against machine learning.
\newblock In {\em Proceedings of the 2017 ACM on Asia conference on computer
  and communications security},  506--519.
\newblock ACM.

\bibitem[\protect\citeauthoryear{Rezaei and Liu}{2019}]{rezaei2019target}
Rezaei, S., and Liu, X.
\newblock 2019.
\newblock A target-agnostic attack on deep models: Exploiting security
  vulnerabilities of transfer learning.
\newblock {\em arXiv:1904.04334}.

\bibitem[\protect\citeauthoryear{Russakovsky \bgroup et al\mbox.\egroup
  }{2015}]{ILSVRC15}
Russakovsky, O.; Deng, J.; Su, H.; Krause, J.; Satheesh, S.; Ma, S.; Huang, Z.;
  Karpathy, A.; Khosla, A.; Bernstein, M.; Berg, A.~C.; and Fei-Fei, L.
\newblock 2015.
\newblock {ImageNet Large Scale Visual Recognition Challenge}.
\newblock {\em International Journal of Computer Vision (IJCV)}
  115(3):211--252.

\bibitem[\protect\citeauthoryear{Sabour \bgroup et al\mbox.\egroup
  }{2016}]{sabour2015adversarial}
Sabour, S.; Cao, Y.; Faghri, F.; and Fleet, D.~J.
\newblock 2016.
\newblock Adversarial manipulation of deep representations.
\newblock {\em ICLR}.

\bibitem[\protect\citeauthoryear{Shafahi \bgroup et al\mbox.\egroup
  }{2018}]{shafahi2018poison}
Shafahi, A.; Huang, W.~R.; Najibi, M.; Suciu, O.; Studer, C.; Dumitras, T.; and
  Goldstein, T.
\newblock 2018.
\newblock Poison frogs! targeted clean-label poisoning attacks on neural
  networks.
\newblock In {\em Advances in Neural Information Processing Systems},
  6103--6113.

\bibitem[\protect\citeauthoryear{Shan \bgroup et al\mbox.\egroup
  }{2019}]{shan2019gotta}
Shan, S.; Willson, E.; Wang, B.; Li, B.; Zheng, H.; and Zhao, B.~Y.
\newblock 2019.
\newblock Gotta catch'em all: Using concealed trapdoors to detect adversarial
  attacks on neural networks.
\newblock {\em arXiv:1904.08554}.

\bibitem[\protect\citeauthoryear{Steinhardt, Koh, and
  Liang}{2017}]{steinhardt2017certified}
Steinhardt, J.; Koh, P. W.~W.; and Liang, P.~S.
\newblock 2017.
\newblock Certified defenses for data poisoning attacks.
\newblock In {\em Advances in neural information processing systems},
  3517--3529.

\bibitem[\protect\citeauthoryear{Suciu \bgroup et al\mbox.\egroup
  }{2018}]{suciu2018does}
Suciu, O.; Marginean, R.; Kaya, Y.; Daume~III, H.; and Dumitras, T.
\newblock 2018.
\newblock When does machine learning $\{$FAIL$\}$? generalized transferability
  for evasion and poisoning attacks.
\newblock In {\em 27th $\{$USENIX$\}$ Security Symposium ($\{$USENIX$\}$
  Security 18)},  1299--1316.

\bibitem[\protect\citeauthoryear{Tran, Li, and Madry}{2018}]{tran2018spectral}
Tran, B.; Li, J.; and Madry, A.
\newblock 2018.
\newblock Spectral signatures in backdoor attacks.
\newblock In {\em Advances in Neural Information Processing Systems},
  8000--8010.

\bibitem[\protect\citeauthoryear{Turner, Tsipras, and
  Madry}{2018}]{turner2018clean}
Turner, A.; Tsipras, D.; and Madry, A.
\newblock 2018.
\newblock Clean-label backdoor attacks.

\bibitem[\protect\citeauthoryear{Wang \bgroup et al\mbox.\egroup
  }{2019}]{wang2019neural}
Wang, B.; Yao, Y.; Shan, S.; Li, H.; Viswanath, B.; Zheng, H.; and Zhao, B.~Y.
\newblock 2019.
\newblock Neural cleanse: Identifying and mitigating backdoor attacks in neural
  networks.
\newblock In {\em IEEE Symposium on Security and Privacy}.

\bibitem[\protect\citeauthoryear{Xiao \bgroup et al\mbox.\egroup
  }{2015}]{xiao2015support}
Xiao, H.; Biggio, B.; Nelson, B.; Xiao, H.; Eckert, C.; and Roli, F.
\newblock 2015.
\newblock Support vector machines under adversarial label contamination.
\newblock {\em Neurocomputing} 160:53--62.

\bibitem[\protect\citeauthoryear{Xiao, Xiao, and
  Eckert}{2012}]{xiao2012adversarial}
Xiao, H.; Xiao, H.; and Eckert, C.
\newblock 2012.
\newblock Adversarial label flips attack on support vector machines.
\newblock In {\em ECAI},  870--875.

\bibitem[\protect\citeauthoryear{Zhu \bgroup et al\mbox.\egroup
  }{2019}]{zhu2019transferable}
Zhu, C.; Huang, W.~R.; Shafahi, A.; Li, H.; Taylor, G.; Studer, C.; and
  Goldstein, T.
\newblock 2019.
\newblock Transferable clean-label poisoning attacks on deep neural nets.
\newblock {\em arXiv:1905.05897}.

\end{thebibliography}

\end{document}